\documentclass[sigplan,screen]{acmart}
\AtBeginDocument{%
  }

\usepackage[ruled, linesnumbered]{algorithm2e}
\usepackage{multirow}

\begin{document}

\copyrightyear{2026}
\acmYear{2026}
\setcopyright{cc}
\setcctype{by}
\acmConference[EuroMLSys '26]{Sixth European Workshop on Machine Learning and Systems }{April 27--30, 2026}{Edinburgh, Scotland Uk}
\acmBooktitle{Sixth European Workshop on Machine Learning and Systems (EuroMLSys '26), April 27--30, 2026, Edinburgh, Scotland Uk}
\acmDOI{10.1145/3805621.3807624}
\acmISBN{979-8-4007-2605-7/2026/04}

\title{Cost-Aware Model Orchestration for LLM-based Systems}

\author{Daria Smirnova}
\email{d.smirnova1@lancaster.ac.uk}
\orcid{XXX}
\affiliation{%
  \institution{Lancaster University}
  \city{Lancaster}
  \country{UK}
}

\author{Hamid Nasiri}
\email{h.nasiri@lancaster.ac.uk}
\orcid{XXX}
\affiliation{%
  \institution{Lancaster University}
  \city{Lancaster}
  \country{UK}
}

\author{Marta Adamska}
\email{m.adamska@lancaster.ac.uk}
\orcid{XXX}
\affiliation{%
  \institution{Lancaster University}
  \city{Lancaster}
  \country{UK}
}

\author{Zhengxin Yu}
\email{z.yu8@lancaster.ac.uk}
\orcid{XXX}
\affiliation{%
  \institution{Lancaster University}
  \city{Lancaster}
  \country{UK}
}

\author{Peter Garraghan}
\email{p.garraghan@lancaster.ac.uk}
\orcid{XXX}
\affiliation{%
  \institution{Lancaster University}
  \city{Lancaster}
  \country{UK}
}


\begin{abstract}
As modern artificial intelligence (AI) systems become more advanced and capable, they can leverage a wide range of tools and models to perform complex tasks. The task of orchestrating these models is increasingly performed by Large Language Models (LLMs) that rely on qualitative descriptions of models for decision-making. However, the descriptions provided to existing LLM-based orchestrators frequently do not reflect true model capabilities and performance characteristics, leading to suboptimal model selection, reduced task accuracy, and increased cost. In this paper, we conduct an empirical analysis of LLM-based orchestration limitations and propose a cost-aware model selection method that accounts for performance–cost trade-offs by incorporating quantitative model performance characteristics within decision-making. Initial experimental results demonstrate that our proposed method increases accuracy by 0.90\%–11.92\% across various evaluated tasks, achieves up to a 54\% energy efficiency improvement, and reduces orchestrator model selection latency from 4.51 s to 7.2 ms.
\end{abstract}



\maketitle

\section{Introduction}
\label{introduction}

AI systems can facilitate many complex tasks by making use of external tools and models that expand system capability. The key component of such AI systems is the orchestrator, responsible for planning, tool selection, and workflow execution to complete tasks \cite{llava-plus}. With the widespread adoption and rapid scaling of such systems, there is an urgent need to design AI system orchestrators that operate at high speed, accuracy, and low cost.

In response to this demand, researchers have identified that due to their reasoning capabilities, LLMs can be instructed to operate as system orchestrators. These \textit{LLM-based orchestrators} use foundational models to reason, make decisions, and act to invoke specific tools and models \cite{hugginggpt, vipergpt, visualchatgpt}.

Despite the aforementioned advantages, existing LLM-based orchestrators encounter a variety of challenges with respect to their performance and efficiency:

(i) \textbf{Orchestrator decision making for model selection is limited to qualitative data of a model's capability}, specifically model cards and textual descriptions such as recommended tasks, examples of usage, or number of user likes on HuggingFace \cite{modelselection}. Such information does not capture task-level performance of these models (i.e. accuracy, cost). Sole reliance on qualitative data results in suboptimal model choices, which undermines the cost-performance trade-offs necessary for system efficiency \cite{modelpopularity}.

(ii) \textbf{LLM-based orchestration requires extensive communication with an LLM to operate} --- textual descriptions of models, such as API descriptions and usage examples of each tool, must be sent to the LLM to provide context \cite{toolformer}. This incurs heavy token usage and latency, which increases cost such as energy consumption, and reduces system throughput. Moreover, the LLM's context window limits the number of tools the orchestrator is able to consider \cite{toolkengpt}.

To address these limitations, we postulate that incorporating quantitative metrics, such as accuracy and cost, into orchestrator decisions can yield more effective model selection and enable the LLM-orchestrated system to balance accuracy and cost-efficiency. In this work, we define cost as the energy consumed per request, reported in Joules (J). We adopt an energy-based cost definition because it is directly measurable and is increasingly important given the growing concerns of AI sustainability and data center growth \cite{energy-footprint}. 

In this work, we present \textbf{an empirical analysis of state-of-the-art LLM-based model orchestration methods}, demonstrating current performance limitations. Through conducting in-depth experimentation of LLM-orchestration across an assortment of different task types, we identify that existing LLM orchestrators often invoke more models than necessary and systematically make suboptimal model choices, leading to higher cost and lower accuracy, while incurring substantial overhead. From these findings, this work then \textbf{explores the feasibility of a quantitative-driven Cost-Aware Model Orchestration framework (CAMO)}. Our proposed approach leverages an online system-state monitoring (energy tracking) that selects models using Pareto optimization to balance performance under an available cost budget (J). Our initial experiments indicate capturing both qualitative and quantitative model information enables more effective orchestration: CAMO achieves higher accuracy and improved energy efficiency while significantly reducing overhead of LLM-based selectors from 4.51 s to 7.2 ms.

\section{Empirical analysis of current LLM-based orchestrators}\label{sec:formulation}

\subsection{Evaluation Objective}
Our objective is to empirically analyse existing LLM-based orchestrators, specifically how they perform model selection, to what extent they consider performance and cost, and what overhead can arise from the current approach.

\begin{table}[t]
\caption{Task types used in our evaluation.}
\label{tab:task-types}
\centering
\small
\setlength{\tabcolsep}{6pt}
\renewcommand{\arraystretch}{1.1}

\begin{tabular}{@{}l
p{0.45\columnwidth}@{} }
\toprule
Abbrev. & Task Type \\
\midrule
\textbf{ICapt} & Image Captioning \\
\textbf{VQA}   & Visual Question Answering \\
\textbf{OD}    & Object Detection\\
\textbf{IGen}  & Image Generation \\
\bottomrule
\end{tabular}
\end{table}

We leverage JARVIS (Hugging GPT), a popular framework that connects an LLM with a multitude of models hosted on HuggingFace via APIs \cite{hugginggpt}, as our selected case study. We evaluated four tasks commonly used in LLM-based systems: ICapt, VQA, OD and IGen (Table \ref{tab:task-types}).  For VQA and OD we used 5,000 samples each from \textit{lmms-lab/OK-VQA} and \textit{COCO2017} datasets on HuggingFace, respectively. For ICapt and IGen we used 1,000 samples each from the \textit{jpawan33/fkr30k} and \textit{Falah/SDXL} datasets, respectively (we used less samples for ICapt and IGen due to lower selection variability).

Each prompt was sent via an HTTP request to the JARVIS server, set up locally on an Ubuntu machine, and equipped with an NVIDIA RTX 6000 Ada GPU. This machine hosted 26 ML models for these four task types (Appendix \ref{app:models-profiling}). The data collected in this experiment includes details on the task types identified for each prompt, the selected models, the corresponding latency, CPU and GPU energy consumption. For CPU and GPU energy data collection, we used Turbostat and Zeus Python libraries, respectively.

\subsection{Analysis Results}
Upon the analysis of the results, two main issues with LLM-orchestrated decision making were discovered:

\textbf{LLMs often selects the wrong task type or decide that more than one task is required}. For example, in 99.2\% of VQA prompts, the LLM built a workflow with at least 2 tasks (Table \ref{table:task-comb}). Whenever an LLM identifies a VQA prompt correctly as a \textit{VQA} task, the average GPU energy consumption is only 65 J (corresponding to 0.3\% of all VQA prompts), whereas the most popular combination \textit{ICapt, VQA} (70\% of prompts) has a mean energy of 224 J. Similarly, correctly identifying an OD prompt can use approximately two times less energy than the predominant \textit{OD, VQA} combination (18.2 J compared to 37.5 J). 

Accuracy follows the same pattern: it is the highest (92.3\%) for VQA prompts that are identified correctly as a \textit{VQA} task, and in the OD experiment, the predominant \textit{OD, VQA} combination achieves 20.1\%, compared to 68.2\% for the correct \textit{OD} combination. This means that using multiple models to answer a user's prompt can paradoxically lead to lower-quality responses, while increasing energy cost. This likely happens because the LLM loses some information as it consolidates model outputs into a single coherent response for the user. This exposes a resource-efficiency issue in LLM-orchestrated systems that must be addressed to improve their efficiency.

\begin{table}[ht]
\footnotesize
\caption{Prompt-level task combinations with associated energy usage and accuracy.}
\label{table:task-comb}
\vskip 0.15in
\centering

\footnotesize
\renewcommand{\arraystretch}{1.0}

\begin{tabular}{l r r r}
\toprule
\multicolumn{1}{l}{\shortstack[l]{Task\\ Combination}} &
\multicolumn{1}{c}{\shortstack[c]{Proportion\\ of Prompts}} &
\multicolumn{1}{c}{\shortstack[c]{Avg.\\Energy (J)}} &
\multicolumn{1}{c}{\shortstack[c]{Accuracy}} \\
\midrule

\addlinespace[4pt]
\multicolumn{4}{c}{\textbf{VQA Dataset}} \\
\addlinespace[2pt]

ICapt, VQA & 70\% & 224 & 62.5\% \\
ICapt, VQA, OD & 17.5\% & 581 & 65.0\% \\
ICapt, DocVQA & 8.3\% & 576 & 22.0\% \\
VQA, IClass & 2.3\% & 129 & 76.8\% \\
ICapt & 0.5\% & 167 & 55.6\% \\
\textbf{VQA} & \textbf{0.3\%} & \textbf{65} & \textbf{92.3\%} \\
ICapt, OD & 0.2\% & 435 & 50.0\% \\
VQA, OD & 0.1\% & 334 & 85.7\% \\

\addlinespace[4pt]
\midrule
\multicolumn{4}{c}{\textbf{OD Dataset}} \\
\addlinespace[2pt]

OD, VQA & 99.2\% & 37.5 & 20.1\% \\
\textbf{OD} & \textbf{0.7\%} & \textbf{18.2} & \textbf{68.2\%} \\
VQA & 0.1\% & 19.4 & 1.7\% \\

\bottomrule
\end{tabular}%

\end{table}

\begin{figure}[t]
    \centering  
    \includegraphics[width=\columnwidth]{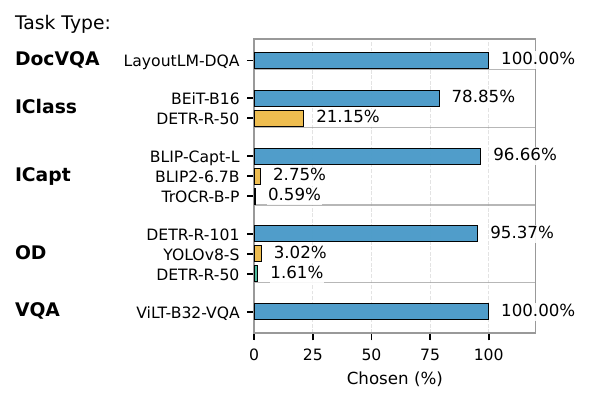}
    \caption{Model selections done by Name-Only on the VQA dataset. DocVQA is Document Visual Question Answering, IClass is Image Classification.}
    \label{fig:name-only-selections}
\end{figure}

\textbf{Even with correct task identification, LLM orchestrators still systematically choose suboptimal models.}
We observed two main issues with LLM-based model selection: (1) a strong bias towards more popular models when model metadata contains popularity metrics, such as HuggingFace model card likes, and (2) overreliance on the LLM’s internalized knowledge about models when such metadata is missing from model descriptions. We substantiate these findings below via selection-frequency distributions across four datasets (Figure~\ref{fig:name-only-selections}, Appendix~\ref{app:model-selection}) and accuracy/energy profiling of all candidates (Appendix~\ref{app:models-profiling}).

\textbf{Popularity-Based Selection Bias.}
Our analysis shows that, in most cases, the LLM favors the model with the greatest recognition on HuggingFace (by the number of likes). 

As an example, 100\% of prompts from the ICapt dataset sent to JARVIS resulted in the LLM choosing the \textit{ViT-GPT2} model out of five candidate ICapt models available on our server. After profiling all candidate models on items from the same dataset (Appendix \ref{app:models-profiling}), we found that \textit{ViT-GPT2} (accuracy: 0.284, energy: 12.7) is neither the most energy-efficient model nor the most accurate among the five candidates. The model with the highest accuracy for the ICapt task is \textit{BLIP2-6.7B} (accuracy: 0.320, energy: 110.8), while the best trade-off between accuracy and cost is achieved by \textit{BLIP-Capt-B} (accuracy: 0.315, energy: 12.5). Despite that, \textit{BLIP-Capt-B} was never selected due to its low number of likes (i.e., 44). 

A similar issue is observed in the VQA dataset experiment. The LLM selected \textit{ViLT-B32-VQA} (accuracy: 0.414, energy: 6.5) for 100\% of all VQA tasks. If the LLM instead selected \textit{BLIP-VQA-B} (accuracy: 0.531, energy: 6.9), accuracy could have increased by 28.3\%, adding only 6.2\% in energy cost.

These results indicate that the number of likes and downloads is not a reliable metric for model quality or efficiency. Therefore, qualitative descriptions should not drive model selection; they should only provide context for understanding model use cases.

\textbf{Opaque model selection in the absence of popularity metrics.} Removing the popularity data from the model metadata did not solve these issues—the LLM still made suboptimal choices. Without the popularity data, the LLM exhibited a strong preference for certain models, driven by its internal knowledge, which can be too general to reflect models' actual performance.

To investigate this further, we conducted an additional set of experiments using the same setup, but excluded the number of likes and downloads from the model descriptions. For clarity, from this point onward, we call the default LLM-based model selection “JARVIS”, and the variant without popularity information as “Name-Only”. Full results of Name-Only experiments are presented in Appendix \ref{app:model-selection}. 

On the OD task, Name-Only showed a strong preference towards \textit{DETR-R-101}, selecting it in 99.82\% of cases during the OD experiment, and in 95.37\% of cases during the VQA experiment (Figure \ref{fig:name-only-selections}), stating in the reasoning that this model has higher performance capabilities than \textit{DETR-R-50} (Appendix \ref{appendix:appendix-OD}). However, this information was not included in the provided model description; therefore, this leads us to the assumption that the LLM relies mostly on LLM's prior internal knowledge about the performance of these models. Hence, relying on LLMs' judgment calls is highly unreliable, as they do not reflect the model's true capabilities and can exhibit biases due to the data they were trained on.

As shown in Figure \ref{fig:name-only-selections}, for ICapt tasks, \textit{BLIP-Capt-L} (accuracy: 0.531, energy: 7.3) is consistently chosen. However, our model profiling shows that \textit{BLIP-Capt-B} (accuracy: 0.531, energy: 6.9) is just as accurate but consumes less energy, yet it was never selected.

These results suggest that LLMs possess a biased internal understanding of ML models, often failing to capture their objective  capabilities. Therefore, when model selection is driven by popularity or LLM's internal knowledge, the resulting performance is consistently suboptimal.
\begin{figure}[t]
  \centering
  \includegraphics[width=\columnwidth]{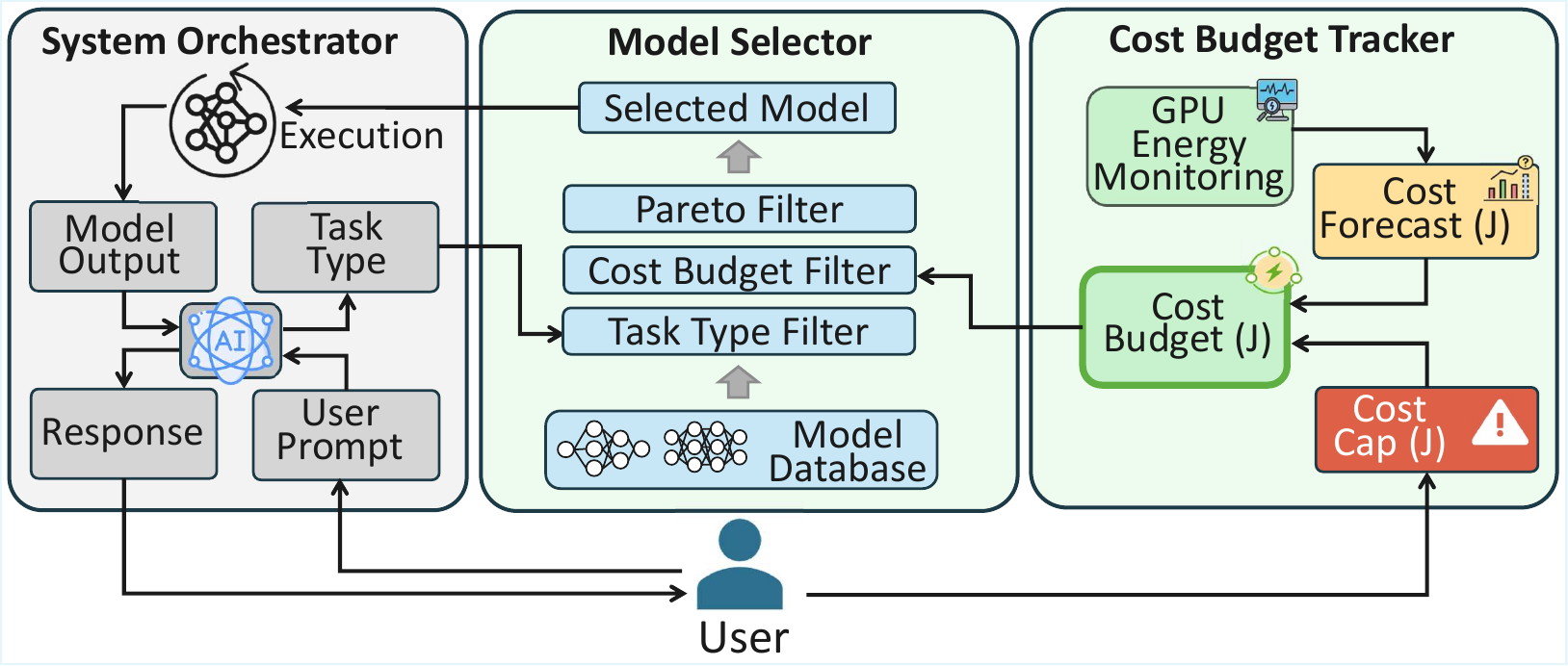}
  \caption{Overview of the cost-aware data-driven model selection framework. Cost is measured as energy usage (J).}
  \label{fig:framework}
  
\end{figure}

\section{Cost-aware Data-driven Model Selection}\label{sec:method}
Based on the analysis findings surfaced in Section 2, we propose that model selection in LLM-orchestrated systems should rely on explicit quantitative performance and cost metrics, such as accuracy and energy consumption. To test the hypothesis, we developed CAMO, a data-driven model selection method, and incorporated it into JARVIS. This allowed for a detailed evaluation of the effects of the proposed model selection framework on the efficiency and performance of LLM-orchestrated systems.

Figure~\ref{fig:framework} shows the end-to-end pipeline, which consists of three main components: (i) the \emph{System Orchestrator}, (ii) the \emph{Model Selector}, and (iii) the \emph{Cost Budget Tracker}. We use JARVIS (HuggingGPT) as the off-the-shelf system orchestrator and do not count it as part of our contribution. In our pipeline, JARVIS (a) performs task identification from the user prompt (e.g., ICapt, VQA, OD, IGen), (b) executes the model(s) selected for that task, and (c) produces the final response to the user. Our contribution is the two remaining components: the cost budget tracker and the model selector, which help ensure the selected models are high-performing and cost-efficient.

Designed for cost-aware orchestration, CAMO constrains model selection using a user-defined energy (cost) cap $C$ in Joules per time slot. Since we define cost as the GPU energy per request, we implement an energy monitor to continuously measure GPU energy usage and use an exponentially weighted moving average (EWMA) to estimate the remaining energy budget for the current time slot.

\paragraph{Cost budget tracker.}\label{sec:tracker}
The cost budget tracker runs in the background and computes the usable energy budget for the current time slot using GPU energy measurements (Appendix \ref{app:tracker}). Given time slot duration $S$, user-defined cap $C$ (J), polling interval $\Delta t$, and EWMA weight $\alpha \in (0,1)$, the tracker forecasts the energy that will be consumed in the current time slot and calculates the remaining per-slot energy available for additional work, such as inference.

\paragraph{Model selector}\label{sec:selector}
The Model Selector first filters the model set $\mathcal{M}$ by task type $\tau$ (Algorithm \ref{alg:selector}). Then, using the usable energy budget  $E_{\text{usable}}$, retrieved from the tracker, it finds only the models that fit within this energy budget ($E_{\text{avg}}(m) \leq E_{\text{usable}}$). The Selector then performs Pareto-Efficient filtering to obtain a subset of models on the Pareto Frontier of $(Acc, E_{\text{avg}})$. From this subset, it selects the model with the highest accuracy on the given task. This process balances accuracy and energy consumption while adhering to the user-defined cap.

\begin{algorithm}[t]

\caption{Model Selector}
\label{alg:selector}
\KwIn{Model set $\mathcal{M}=\{(m,\tau(m),E_{\text{avg}}(m),Acc(m))\}$;\\
usable energy budget $E_{\text{usable}}$; task type $\tau$; $\Delta t_{\text{retry}}$}
\KwOut{Selected model $m^*$}

\Repeat{$|\mathcal{M}_{\text{budget}}| > 0$}{
    $\mathcal{M}_\tau \leftarrow \{\, m \in \mathcal{M} \mid \tau(m) = \tau \,\}$\;
    $E_{\text{usable}} \leftarrow \text{Tracker.pull()}$
    $\mathcal{M}_{\text{budget}} \leftarrow \{\, m \in \mathcal{M}_\tau \mid E_{\text{avg}}(m) \leq E_{\text{usable}} \,\}$\;
    \If{$|\mathcal{M}_{\text{budget}}| = 0$}{
        \tcp{Wait $\Delta t_{\text{retry}}$ seconds}
        \lIf{$\Delta t_{\text{retry}} > 0$}{\text{sleep}($\Delta t_{\text{retry}}$)}
    }
}
$\mathcal{P} \leftarrow \{\, m \in \mathcal{M}_{\text{budget}} \mid \nexists\, m' \in \mathcal{M}_{\text{budget}} : E_{\text{avg}}(m') < E_{\text{avg}}(m) \wedge Acc(m') > Acc(m) \,\}$\;
$m^* \leftarrow \arg\max_{m \in \mathcal{P}} Acc(m)$\;

\end{algorithm}

\section{Performance Evaluation}\label{sec:experiments}

\subsection{Experiment Setup} We evaluate our proposed cost-aware model selection framework by evaluating its effectiveness in comparison to existing prominent model selection methods.

\textbf{Hardware.} All experiments ran on Ubuntu 20.04.6 LTS using a single NVIDIA RTX 6000 Ada GPU (50 GB VRAM), with NVIDIA driver 535.247.01 (CUDA 12.2), on an Intel Xeon Gold 5418Y host.

\textbf{Software.} We perform the evaluation within the JARVIS framework. The model selector and the cost budget tracker are implemented in Python. The cost budget tracker is a separate, parallel-running process, while the model selector that communicates with the tracker makes per-request model choices. Inference is executed by the framework's built-in execution engine.

\textbf{Model Selection Methods.} 
We compare three model selection policies within the identical JARVIS pipeline:

\begin{itemize}
    \item \textit{JARVIS}: The framework's default method that prioritizes models with the highest number of likes.
    \item \textit{Name-Only}: The knowledge-based method that makes decisions based on textual model descriptions.
    \item \textit{CAMO}: Our proposed method with four user-defined cost targets (100 J, 150 J, 400 J, and 600 J). System names use a suffix to denote the cost cap (e.g., CAMO-100 = cost cap of 100 J).
\end{itemize}

\textbf{Models.}
 We use the 26 models described in Section \ref{sec:formulation} analysis (Appendix \ref{app:models-profiling}). The default built-in execution engine of JARVIS loads and unloads a model for every request; therefore, we changed this logic to keep models in memory for the whole experiment to reflect real-world system dynamics. JARVIS and Name-Only use GPT-4o-mini as the model orchestrator, chosen for its high performance and affordable cost.

\textbf{Datasets.}
We evaluate all three policies using 100-prompt subsets from the datasets specified in Section~\ref{sec:formulation}. For each policy and dataset, we execute one end-to-end run per prompt and report aggregate statistics across prompts. Note that each prompt may result in multiple tasks and, therefore, multiple model executions.

\textbf{Evaluation metrics.} 
ROUGE-L score is used to assess the quality of models' performance on VQA and ICapt tasks. To evaluate the performance of OD models, we use the standard mAP@0.5 metric. To report the response quality of IGen models, we use CLIP (openai/clip-vit-base-patch32). For energy usage, we use the standard unit of Joules (J). We also report Accuracy per Joule (Acc/J) as an energy-efficiency metric that captures the accuracy achieved per unit of energy consumed which allows for a fair comparison of model selections across different policies \cite{acc-j2,acc-j1}.


\subsection{Experiment Results and Analysis}
\begin{figure}[!t]
    \centering
    
    \includegraphics[width=\columnwidth]{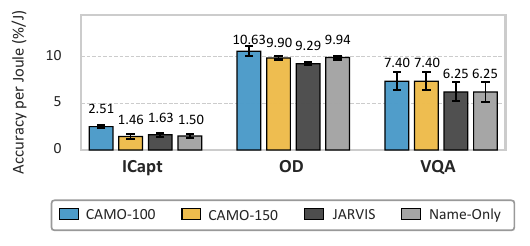}
    \caption{Accuracy per Joule, calculated on weighted results (ICapt, VQA and OD task types). Our proposed method CAMO (with two cost caps of 100 J and 150 J) and two baselines. Error bars represent 95\% CI.}
    \label{fig:eval_results1}
\end{figure}

\textbf{Relying on qualitative descriptions in model selection degrades system accuracy and cost efficiency}. Table \ref{tab:main_results} shows that model selection based on quantitative data improves accuracy across all four tasks (absolute gains from 0.90\% to 11.92\%). The accuracy of our proposed method is higher than the baselines on ICapt and VQA tasks. Moreover, for the ICapt, VQA, and OD tasks, it achieves better accuracy-energy trade-offs (Figure \ref{fig:eval_results1}). CAMO-100 demonstrates the largest efficiency gains, improving Accuracy-per-Joule by 54\%, 14.4\%, and 18.4\% over JARVIS on ICapt, OD, and VQA, respectively. It also outperforms Name-Only system on this metric by 67\% on ICapt, 6.9\% on OD, and 18.4\% on VQA. The accuracy improvement in IGen can be attributed to a more relaxed energy level of 600 J, allowing for larger models.

\begin{table}[t]
\centering
\small
\caption{Systems comparison by task, aggregated over requests from four datasets. Sorted by Acc/J.}
\vskip 0.10in

\begin{tabular}{l l r r r}
\toprule
Task & System & Acc (\%) & Energy (J) & Acc/J (\%/J) \\
\midrule

\multirow{4}{*}{\textbf{ICapt}} 
 & CAMO-100        & 31.50 & 12.54 & \textbf{2.51} \\
& Jarvis            & 28.60 & 17.58 & 1.63 \\
& Name-Only & 28.67 & 19.07 & 1.50 \\
& CAMO-150         & \textbf{31.55} & 21.68 & 1.46 \\

\midrule
\multirow{4}{*}{\textbf{OD}} 
& CAMO-100         & 70.03 &  6.59 & \textbf{10.63} \\
& Name-Only & 71.11 &  7.15 & 9.94 \\
& CAMO-150         & \textbf{71.26} &  7.20 & 9.90 \\
& Jarvis            & 65.00 &  7.00 & 9.29 \\

\midrule
\multirow{4}{*}{\textbf{VQA}} 
& CAMO-150         & \textbf{52.54} & 7.10 & \textbf{7.40} \\
& CAMO-100         & 52.50 & 7.09 & \textbf{7.40} \\
& Jarvis            & 40.62 & 6.50 & 6.25 \\
& Name-Only & 40.62 & 6.50 & 6.25 \\

\midrule
\multirow{4}{*}{\textbf{IGen}} 
& CAMO-400         & 34.20 & 193.33 & \textbf{0.18} \\
& Jarvis            & 34.20 & 193.33 & \textbf{0.18} \\
& Name-Only & 34.19 & 193.40 & \textbf{0.18} \\
& CAMO-600       & \textbf{35.10} & 431.64 & 0.08 \\
\bottomrule
\end{tabular}
\label{tab:main_results}
\end{table}

\begin{figure}[htbp]
    \centering
    \includegraphics[width=\columnwidth]{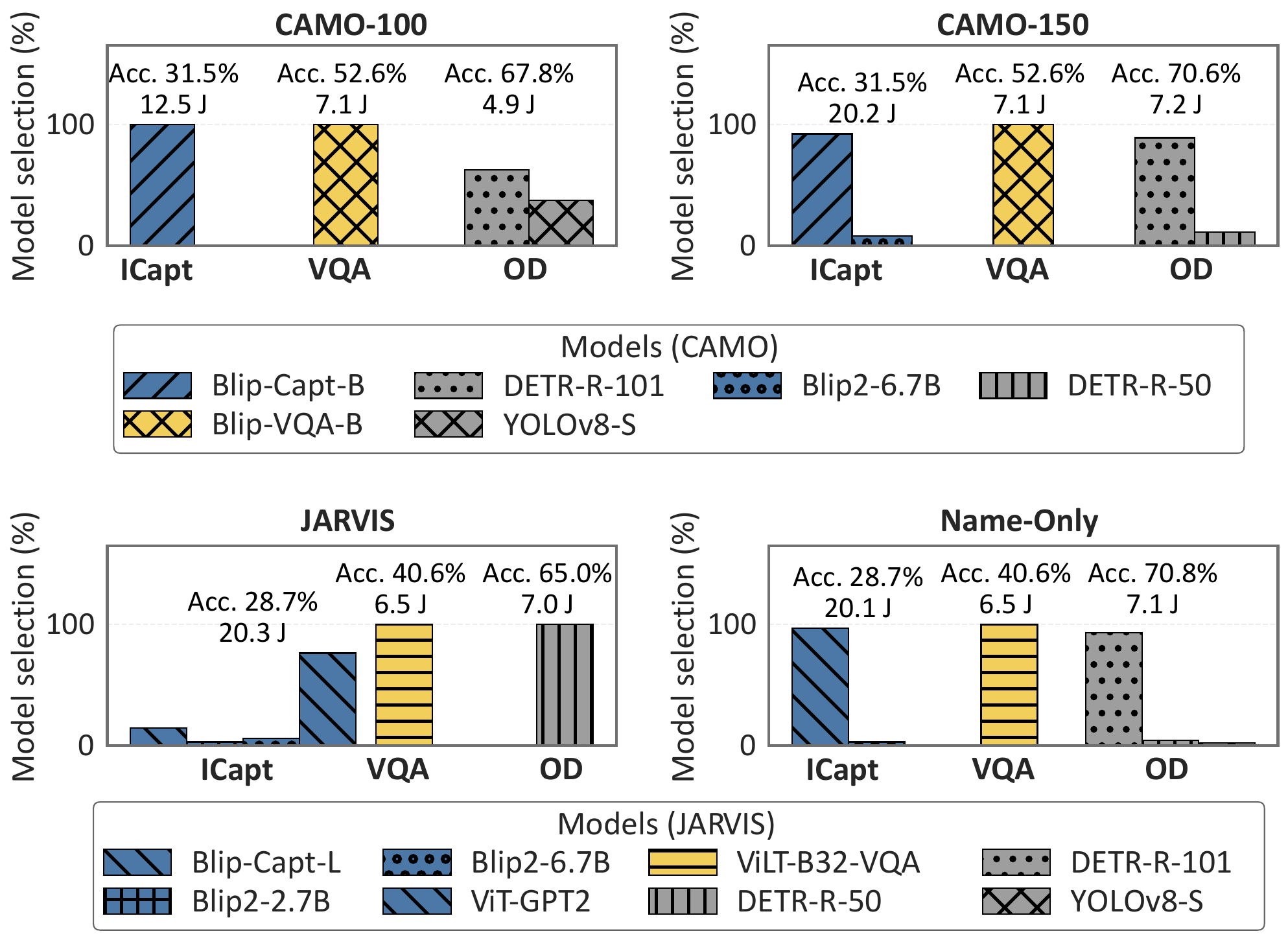}
    \caption{Model selection performance on the VQA dataset.}
    \label{fig:vqa_eval}
\end{figure}

\textbf{Data-driven model selection enables Pareto-efficient selections, leading to a better balance between performance and cost.} Our method selects 100\% of models on the Pareto frontier of (accuracy, energy). In contrast, JARVIS and Name-Only systems selected Pareto-efficient models in 72.3\% and 72.7\% of cases, respectively. Hence, in at least 28\% of all model selections, there existed at least one other model that was better in both accuracy and energy consumption. Our model selection approach avoids making such choices, ensuring a more efficient use of available resources.

\textbf{CAMO operates at a fraction of the overhead, compared to the LLM-based model selectors.}
CAMO introduces a mean latency of 7.2 ms per request and consumes on average 21.9 mJ on the CPU per decision (from 10,000 requests). The NVML-based energy tracker contributes on average 1 mJ of CPU energy per selection and, due to constant polling of GPU readings (NVML queries) at 100 ms intervals, adds 0.316 W overhead to GPU power consumption (about 18.96 J per minute). For the LLM-based selectors (based on 10,000 selections using JARVIS), the selection step uses on average 2,081 tokens per request (input: 1,977, output: 104) and incurs a mean selection time of 4.51 s. For each decision, JARVIS provides in-context examples and a list of candidate models, resulting in a large number of tokens and, consequently, high latency, leading to impaired quality of service (QoS).

Overall, CAMO achieves \(6.4 \times 10^{2}\) lower latency (4.51 s vs. 7.2 ms) and negligible energy overhead, in comparison to the LLM-based methods.

\begin{figure}[t]
    \centering
    \includegraphics[width=\columnwidth]{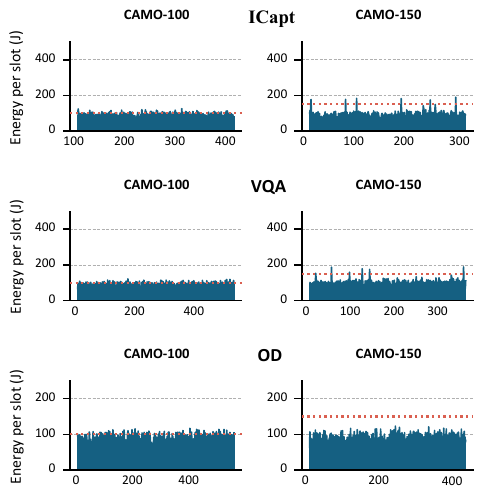}
    \caption{Per-slot energy usage. The dotted red line indicates the cost cap; blue bars --- realized energy usage. Includes GPU base draw of 40-50 J/s. The x-axis denotes time slots.}
    \label{fig:violations}
\end{figure}

\textbf{The proposed method adheres closely to user-defined cost caps}, especially at caps $\geq$ 150 J (Figure \ref{fig:violations}). Some overshoots occur at the 100 J cap because the target is close to baseline energy variability, reducing the prediction stability. 

\textbf{In the absence of popularity signals, an LLM-based model selector outperforms a popularity-driven baseline on OD tasks.} While on the VQA task the two baselines perform comparably, the OD data indicate a clear advantage for Name-Only (Figure \ref{fig:eval_results1}). Compared with JARVIS, it achieves 6.11\% higher accuracy with only 0.15 J higher energy cost (Table \ref{tab:main_results}) and higher Accuracy-per-Joule (+0.65\%/J). The differences are insignificant across the other three tasks. This confirms that popularity metadata in model descriptions can degrade system performance.

\textbf{CAMO replaces LLM-driven selection with a local, data-driven policy that boosts accuracy, cuts costs, and avoids both token usage and high latency.}

\section{Related Work}\label{sec:related_work}
Inefficient use of LLMs and the costs they incur have been extensively studied across many domains. For example, modern LLM-serving systems can switch between model variants based on query complexity, utilizing larger LLMs only when necessary, minimizing costs \cite{tryage,hybridllm,routellm}. Beyond model selection, specialized LLM-serving scheduling techniques have been proposed to ensure system efficiency \cite{thunderserve, demystifying, greenllm}. However, these approaches focus on LLM serving, not orchestration policies that select among many tools and heterogeneous ML models using external signals.

As for cost-aware operation, prior work has explored energy- and cost-aware scheduling in large-scale systems. For example, data centers commonly schedule workloads under energy constraints \cite{pcaps}, while energy-aware scheduling is commonly used in model-serving systems \cite{infaas,m-serve}. Within LLM pipelines, cost-aware tool selection has also been previously explored \cite{catp-llm}, and related resource-aware adaptation appears in IoT and other constrained settings \cite{sabovic,bullo}. Yet, none of those systems use real-time energy data with quantitative model performance metrics when selecting among heterogeneous models in LLM-orchestrated systems. The proposed method closes this gap by making cost-aware data-driven selections in the orchestration layer.

\section{Limitations and Future Work}
Our results show clear advantages of cost-aware data-driven selection policies over black-box, LLM-based decision-making, as they achieve better cost-performance trade-offs while significantly reducing overhead.

We evaluated four vision tasks on a single hardware setup and define cost as GPU energy per request. While heterogeneous hardware will not affect the accuracy of the given model, it will however alter its energy use, and thus extending this to broader deployment settings and cost definitions (e.g., memory usage) is an important next step. Moreover, our current implementation uses a fixed set of models with offline profiling. In future work, we will move to online per-model profiling so cost and performance data stay up to date as models, hardware and workloads change. We will also extend our method to learn from past data to continuously update its estimates and penalize under-performing models. Moreover, current implementation of CAMO relies on metrics of accuracy and energy that are not conditioned on input size or prompt difficulty. In future work, we will stratify these metrics to estimate cost and performance profiles conditioned on request characteristics (e.g., image resolution, prompt length).

Finally, we plan to support more comprehensive user-defined QoS constraints, such as minimum accuracy or maximum latency, in addition to the cost budget.

\section{Conclusion}\label{sec:conclusion}
In this work, we have analyzed LLM-based model selection methods and their limitations. Our results show that these methods often invoke more models than necessary, misclassify tasks, and make suboptimal model choices. To address these shortcomings, we propose a cost-aware data-driven model selection method, CAMO, that considers real-time energy levels and accuracy metrics and uses Pareto optimization to enable explicit performance-cost trade-offs. Our method achieves significant accuracy improvements on four vision tasks (with absolute gains ranging from 0.90\% to 11.92\%) and up to 54\% in Accuracy-per-Joule on three tasks, while operating at 7.2 ms latency per request (vs 4.51 s for LLM-based methods). These results show that cost-aware data-driven model selection improves both accuracy and cost-efficiency while reducing overhead, making it a promising direction for sustainable and efficient system operation.

\section{Acknowledgments}
This work was supported by the Engineering and Physical Sciences Research Council (Fellowship number EP/V007092/1).

\bibliographystyle{ACM-Reference-Format}
\bibliography{reference}

@misc{hugginggpt,
      title={{HuggingGPT}: Solving {AI} Tasks with {ChatGPT} and its Friends in {Hugging Face}}, 
      author={Yongliang Shen and Kaitao Song and Xu Tan and Dongsheng Li and Weiming Lu and Yueting Zhuang},
      year={2023},
      eprint={2303.17580},
      archivePrefix={arXiv},
      primaryClass={cs.CL},
      url={https://arxiv.org/abs/2303.17580}, 
}

@misc{visualchatgpt,
      title={{Visual ChatGPT}: Talking, Drawing and Editing with Visual Foundation Models}, 
      author={Chenfei Wu and Shengming Yin and Weizhen Qi and Xiaodong Wang and Zecheng Tang and Nan Duan},
      year={2023},
      eprint={2303.04671},
      archivePrefix={arXiv},
      primaryClass={cs.CV},
      url={https://arxiv.org/abs/2303.04671}, 
}

@misc{vipergpt,
      title={{ViperGPT}: Visual Inference via Python Execution for Reasoning}, 
      author={Dídac Surís and Sachit Menon and Carl Vondrick},
      year={2023},
      eprint={2303.08128},
      archivePrefix={arXiv},
      primaryClass={cs.CV},
      url={https://arxiv.org/abs/2303.08128}, 
}

@misc{modelselection,
      title={Towards Robust Multi-Modal Reasoning via Model Selection}, 
      author={Xiangyan Liu and Rongxue Li and Wei Ji and Tao Lin},
      year={2024},
      eprint={2310.08446},
      journal={arXiv preprint arXiv:2310.08446},
    archivePrefix={arXiv},
      primaryClass={cs.LG},
      url={https://arxiv.org/abs/2310.08446}, 
}

@misc{toolformer,
      title={Toolformer: Language Models Can Teach Themselves to Use Tools}, 
      author={Timo Schick and Jane Dwivedi-Yu and Roberto Dessì and Roberta Raileanu and Maria Lomeli and Luke Zettlemoyer and Nicola Cancedda and Thomas Scialom},
      year={2023},
      eprint={2302.04761},
      archivePrefix={arXiv},
      primaryClass={cs.CL},
      url={https://arxiv.org/abs/2302.04761}, 
}

@article{sabovic,
title = {Energy-aware {tinyML} model selection on zero energy devices},
journal = {Internet of Things},
volume = {30},
pages = {101488},
year = {2025},
doi = {10.1016/j.iot.2025.101488},
author = {Adnan Sabovic and Jaron Fontaine and Eli De Poorter and Jeroen Famaey}
}

@misc{bullo,
      title={Energy-Aware Dynamic Neural Inference}, 
      author={Marcello Bullo and Seifallah Jardak and Pietro Carnelli and Deniz Gündüz},
      year={2024},
      eprint={2411.02471},
      archivePrefix={arXiv},
      primaryClass={cs.LG},
      url={https://arxiv.org/abs/2411.02471}, 
}

@misc{greenllm,
      title={{GreenLLM}: {SLO-Aware} Dynamic Frequency Scaling for Energy-Efficient {LLM} Serving}, 
      author={Qunyou Liu and Darong Huang and Marina Zapater and David Atienza},
      year={2025},
      eprint={2508.16449},
      archivePrefix={arXiv},
      primaryClass={cs.PF},
      url={https://arxiv.org/abs/2508.16449}, 
}

@misc{hybridllm,
      title={{Hybrid LLM}: Cost-Efficient and Quality-Aware Query Routing}, 
      author={Dujian Ding and Ankur Mallick and Chi Wang and Robert Sim and Subhabrata Mukherjee and Victor Ruhle and Laks V. S. Lakshmanan and Ahmed Hassan Awadallah},
      year={2024},
      eprint={2404.14618},
      archivePrefix={arXiv},
      primaryClass={cs.LG},
      url={https://arxiv.org/abs/2404.14618}, 
}

@misc{routellm,
      title={{RouteLLM}: Learning to Route {LLMs} with Preference Data}, 
      author={Isaac Ong and Amjad Almahairi and Vincent Wu and Wei-Lin Chiang and Tianhao Wu and Joseph E. Gonzalez and M Waleed Kadous and Ion Stoica},
      year={2025},
      eprint={2406.18665},
      archivePrefix={arXiv},
      primaryClass={cs.LG},
      url={https://arxiv.org/abs/2406.18665}, 
}

@inproceedings{infaas,
  author={Francisco Romero and Qian Li and Neeraja J. Yadwadkar and Christos Kozyrakis},
  title={{INFaaS}: Automated Model-less Inference Serving},
  year={2021},
  pages={397-411},
  booktitle={Proceedings of the 2021 USENIX Annual Technical Conference (USENIX ATC 2021)},
}

@inproceedings{m-serve,
author = {Qiu, Haoran and Mao, Weichao and Patke, Archit and Cui, Shengkun and Jha, Saurabh and Wang, Chen and Franke, Hubertus and Kalbarczyk, Zbigniew T. and Ba\c{s}ar, Tamer and Iyer, Ravishankar K.},
title = {Power-aware deep learning model serving with µ-serve},
year = {2024},
address = {Santa Clara, USA},
booktitle = {Proceedings of the 2024 USENIX Annual Technical Conference (USENIX ATC 2024)},
articleno = {5},
numpages = {19},
pages={75--93},
location = {Santa Clara, CA, USA}
}

@inproceedings{pcaps,
author = {Lechowicz, Adam and Shenoy, Rohan and Bashir, Noman and Hajiesmaili, Mohammad and Wierman, Adam and Delimitrou, Christina},
title = {Carbon- and Precedence-Aware Scheduling for Data Processing Clusters},
year = {2025},
address = {New York, USA},
doi = {10.1145/3718958.3750478},
booktitle = {Proceedings of the ACM SIGCOMM 2025 Conference},
pages = {1241–1244},
numpages = {4},
location = {S\~{a}o Francisco Convent, Coimbra, Portugal}
}

@inproceedings{
acc-j1,
title={The Hidden Joules: Evaluating the Energy Consumption of Vision Backbones for Progress Towards More Efficient Model Inference},
author={Zeyu Yang and Wesley Armour},
booktitle={Forty-second International Conference on Machine Learning},
year={2025},
url={https://openreview.net/forum?id=Va5jZARDcx}
}

@misc{acc-j2,
      title={Towards energy-efficient Deep Learning: An overview of energy-efficient approaches along the Deep Learning Lifecycle}, 
      author={Vanessa Mehlin and Sigurd Schacht and Carsten Lanquillon},
      year={2023},
      eprint={2303.01980},
      archivePrefix={arXiv},
      primaryClass={cs.LG},
      url={https://arxiv.org/abs/2303.01980}, 
}

@misc{tryage,
      title={Tryage: Real-time, intelligent Routing of User Prompts to Large Language Models}, 
      author={Surya Narayanan Hari and Matt Thomson},
      year={2023},
      eprint={2308.11601},
      archivePrefix={arXiv},
      primaryClass={cs.LG},
      url={https://arxiv.org/abs/2308.11601}, 
}

@misc{catp-llm,
      title={{CATP-LLM}: Empowering Large Language Models for Cost-Aware Tool Planning}, 
      author={Duo Wu and Jinghe Wang and Yuan Meng and Yanning Zhang and Le Sun and Zhi Wang},
      year={2025},
      eprint={2411.16313},
      archivePrefix={arXiv},
      primaryClass={cs.AI},
      url={https://arxiv.org/abs/2411.16313}, 
}

@article{od,
author={Vijayakumar,Ajantha and Vairavasundaram,Subramaniyaswamy},
year={2024},
month={10},
title={{YOLO}-based Object Detection Models: A Review and its Applications},
journal={Multimedia Tools and Applications},
volume={83},
number={35},
pages={83535-83574},
isbn={13807501},
language={English},
doi = {10.1007/s11042-024-18872-y}
}

@misc{llava-plus,
      title={{LLaVA-Plus}: Learning to Use Tools for Creating Multimodal Agents}, 
      author={Shilong Liu and Hao Cheng and Haotian Liu and Hao Zhang and Feng Li and Tianhe Ren and Xueyan Zou and Jianwei Yang and Hang Su and Jun Zhu and Lei Zhang and Jianfeng Gao and Chunyuan Li},
      year={2023},
      eprint={2311.05437},
      archivePrefix={arXiv},
      primaryClass={cs.CV},
      url={https://arxiv.org/abs/2311.05437}, 
}

@article{energy-footprint,
title = {The growing energy footprint of artificial intelligence},
journal = {Joule},
volume = {7},
number = {10},
pages = {2191-2194},
year = {2023},
issn = {2542-4351},
doi = {10.1016/j.joule.2023.09.004},
author = {De Vries, Alex},
}

@misc{demystifying,
      title={Demystifying Cost-Efficiency in LLM Serving over Heterogeneous GPUs}, 
      author={Youhe Jiang and Fangcheng Fu and Xiaozhe Yao and Guoliang He and Xupeng Miao and Ana Klimovic and Bin Cui and Binhang Yuan and Eiko Yoneki},
      year={2025},
      eprint={2502.00722},
      archivePrefix={arXiv},
      primaryClass={cs.DC},
      url={https://arxiv.org/abs/2502.00722}, 
}

@misc{thunderserve,
      title={ThunderServe: High-performance and Cost-efficient LLM Serving in Cloud Environments}, 
      author={Youhe Jiang and Fangcheng Fu and Xiaozhe Yao and Taiyi Wang and Bin Cui and Ana Klimovic and Eiko Yoneki},
      year={2025},
      eprint={2502.09334},
      archivePrefix={arXiv},
      primaryClass={cs.DC},
      url={https://arxiv.org/abs/2502.09334}, 
}

@misc{modelpopularity,
      title={Model Hubs and Beyond: Analyzing Model Popularity, Performance, and Documentation}, 
      author={Pritam Kadasi and Sriman Reddy Kondam and Srivathsa Vamsi Chaturvedula and Rudranshu Sen and Agnish Saha and Soumavo Sikdar and Sayani Sarkar and Suhani Mittal and Rohit Jindal and Mayank Singh},
      year={2025},
      eprint={2503.15222},
      archivePrefix={arXiv},
      primaryClass={cs.CL},
      url={https://arxiv.org/abs/2503.15222}, 
}

@misc{toolkengpt,
      title={ToolkenGPT: Augmenting Frozen Language Models with Massive Tools via Tool Embeddings}, 
      author={Shibo Hao and Tianyang Liu and Zhen Wang and Zhiting Hu},
      year={2024},
      eprint={2305.11554},
      archivePrefix={arXiv},
      primaryClass={cs.CL},
      url={https://arxiv.org/abs/2305.11554}, 
}

\appendix

\clearpage
\section{Weighted performance of systems by task (sorted by Acc/J)} \label{appendix:total_results_100}

\begin{table}[h]
\small
\centering
\begin{tabular}{l l r l r l r l}
\toprule
Task & System & Accuracy (\%) & Acc 95\% CI (\%) & Energy (J) & Energy 95\% CI (J) & Acc/J (\%/J) & Acc/J 95\% CI \\
\midrule

\multirow{4}{*}{\textbf{ICapt}}
& CAMO-100         & 31.50 & [28.95, 34.05] & 12.54 & [11.94, 13.14]   & 2.51 & [2.28, 2.75] \\
& Jarvis            & 28.60 & [26.82, 30.38] & 17.58 & [15.15, 20.01]   & 1.63 & [1.38, 1.87] \\
& Name-Only & 28.67 & [26.96, 30.38] & 19.07 & [17.18, 20.96]   & 1.50 & [1.33, 1.68] \\
& CAMO-150         & 31.55 & [30.02, 33.07] & 21.68 & [18.22, 25.14]   & 1.46 & [1.21, 1.70] \\
\addlinespace
\midrule
\multirow{4}{*}{\textbf{VQA}}
& CAMO-150         & 52.54 & [45.75, 59.32] & 7.10 & [7.00, 7.19] & 7.40 & [6.44, 8.37] \\
& CAMO-100         & 52.50 & [45.69, 59.30] & 7.09 & [7.00, 7.19] & 7.40 & [6.44, 8.36] \\
& Jarvis            & 40.62 & [33.91, 47.33] & 6.50 & [6.46, 6.54] & 6.25 & [5.22, 7.28] \\
& Name-Only & 40.62 & [33.82, 47.42] & 6.50 & [6.46, 6.54] & 6.25 & [5.20, 7.30] \\
\addlinespace
\midrule
\multirow{4}{*}{\textbf{OD}}
& CAMO-100         & 70.03 & [69.37, 70.69] &  6.59  & [6.26, 6.92]      & 10.63 & [10.09, 11.17] \\
& Name-Only & 71.11 & [70.67, 71.54] &  7.15  & [7.02, 7.29]      &  9.94 & [9.74, 10.14] \\
& CAMO-150         & 71.26 & [70.84, 71.68] &  7.20  & [7.08, 7.32]      &  9.90 & [9.73, 10.07] \\
& Jarvis            & 65.00 & [64.65, 65.35] &  7.00  & [6.91, 7.09]      &  9.29 & [9.15, 9.42] \\
\addlinespace
\midrule

\multirow{4}{*}{\textbf{IGen}}
& CAMO-400         & 34.20 & [33.47, 34.93] & 193.33 & [190.87, 195.79] & 0.18 & [0.17, 0.18] \\
& Jarvis            & 34.20 & [33.47, 34.93] & 193.33 & [190.89, 195.77] & 0.18 & [0.17, 0.18] \\
& Name-Only & 34.19 & [33.46, 34.92] & 193.40 & [190.96, 195.84] & 0.18 & [0.17, 0.18] \\
& CAMO-600         & 35.10 & [34.43, 35.77] & 431.64 & [429.50, 433.78] & 0.08 & [0.08, 0.08] \\

\bottomrule
\end{tabular}
\end{table}

\section{Model Profiling for Forward Pass and Full Inference Lifecycle} \label{app:models-profiling}
E - Energy, P - Power, T - Time
\begin{table}[h]
\scriptsize
\centering
\resizebox{\textwidth}{!}{%
\begin{tabular}{llllllllllll}
\hline
Task & Model ID & Acc Metric & Acc & E\_fwd & E\_life & T\_fwd & T\_life & P\_fwd & P\_life & GFLOPs \\
\hline

\multirow{5}{*}{\textbf{ICapt}}
& nlpconnect/vit-gpt2-image-captioning     & ROUGE-L & 0.284  & 12.66 & 25.30 & 0.070 & 0.302 & 180.8571 & 83.7748 & 20.07 \\
& Salesforce/blip-image-captioning-base    & ROUGE-L & 0.315  & 12.54 & 27.00 & 0.090 & 0.351 & 139.3333 & 76.9231 & 64.38 \\
& Salesforce/blip-image-captioning-large   & ROUGE-L & 0.286  & 17.23 & 53.50 & 0.100 & 0.600 & 172.3000 & 89.1667 & 202.77 \\
& Salesforce/blip2-opt-2.7b                & ROUGE-L & 0.301  & 43.17 & 319.20& 0.190 & 3.407 & 227.2105 & 93.6895 & 362.21 \\
& Salesforce/blip2-opt-6.7b                & ROUGE-L & 0.320  & 110.84& 701.20& 0.410 & 7.472 & 270.3415 & 93.8437 & 506.37 \\
\midrule

\multirow{9}{*}{\textbf{VQA}}
& dandelin/vilt-b32-finetuned-vqa          & ROUGE-L & 0.4062 & 6.50  & 9.00  & 0.0141& 0.1525& 460.9929 & 59.0164 & 13.79 \\
& Salesforce/blip-vqa-base                 & ROUGE-L & 0.5262 & 7.10  & 30.40 & 0.0319& 0.4465& 222.5705 & 68.0851 & 55.54 \\
& Salesforce/blip-vqa-capfilt-large        & ROUGE-L & 0.5262 & 7.40  & 30.50 & 0.0416& 0.4436& 177.8846 & 68.7556 & 55.54 \\
& microsoft/git-large-vqav2                & ROUGE-L & 0.0561 & 9.50  & 35.30 & 0.0461& 0.4309& 206.0738 & 81.9216 & 394.86 \\
& ivelin/donut-refexp-combined-v1          & ROUGE-L & 0.0406 & 12.60 & 31.90 & 0.0774& 0.3349& 162.7907 & 95.2523 & 335.09 \\
& microsoft/git-base-textvqa               & ROUGE-L & 0.0565 & 8.30  & 14.40 & 0.0343& 0.2169& 241.9825 & 66.3900 & 177.16 \\
& microsoft/git-base-vqav2                 & ROUGE-L & 0.0727 & 7.40  & 14.20 & 0.0325& 0.2157& 227.6923 & 65.8322 & 177.16 \\
& microsoft/git-large-textvqa              & ROUGE-L & 0.0527 & 8.30  & 34.20 & 0.0396& 0.4519& 209.5960 & 75.6805 & 394.86 \\
& tufa15nik/vilt-finetuned-vqasi           & ROUGE-L & 0.4062 & 6.60  & 7.50  & 0.0143& 0.1415& 461.5385 & 53.0035 & 13.79 \\

\midrule
\multirow{6}{*}{\textbf{OD}}
& facebook/detr-resnet-101                 & mAP@0.5 & 0.713  & 7.20  & 15.60 & 0.0360& 0.2021& 220.1258 & 76.8453 & 107.53 \\
& facebook/detr-resnet-50                  & mAP@0.5 & 0.650  & 7.00  & 14.00 & 0.0293& 0.1582& 280.9917 & 94.2998 & 60.18 \\
& hustvl/yolos-tiny                        & mAP@0.5 & 0.5578 & 3.80  & 3.80  & 0.0123& 0.0532& 308.9431 & 71.4286 & 6.17 \\
& hustvl/yolos-small                       & mAP@0.5 & 0.6415 & 7.70  & 15.40 & 0.0500& 0.1564& 154.0000 & 98.4655 & 56.15 \\
& google/owlvit-base-patch32               & mAP@0.5 & 0.5478 & 7.10  & 21.60 & 0.0353& 0.2852& 201.1331 & 75.7363 & 58.69 \\
& ultralyticsplus/yolov8s                  & mAP@0.5 & 0.6199 & 1.20  & 1.20  & 0.0115& 0.0179& 104.3478 & 67.0391 & 14.30 \\

\midrule
\multirow{6}{*}{\textbf{IGen}}
& CompVis/stable-diffusion-v1-4            & CLIP    & 0.338  & 186.72& 495.59& 0.670 & 2.5688& 278.6866 & 192.9248& 339.01 \\
& runwayml/stable-diffusion-v1-5           & CLIP    & 0.342  & 193.33& 510.51& 0.680 & 2.5645& 284.3088 & 199.0689& 339.01 \\
& prompthero/openjourney                   & CLIP    & 0.331  & 200.70& 511.74& 0.690 & 2.5648& 290.8696 & 199.5236& 339.01 \\
& hakurei/waifu-diffusion                  & CLIP    & 0.273  & 193.53& 504.37& 0.670 & 2.6586& 288.8507 & 189.7123& 339.51 \\
& stabilityai/stable-diffusion-2-1         & CLIP    & 0.351  & 431.64& 928.08& 1.470 & 3.7685& 293.6327 & 246.2736& 339.51 \\
& stabilityai/stable-diffusion-2           & CLIP    & 0.348  & 439.27& 938.69& 1.490 & 3.8068& 294.8121 & 246.5820& 339.51 \\
\hline
\end{tabular}
    }
\end{table}
\clearpage

\section{Cost Budget Tracker}
\label{app:tracker}

\begin{algorithm}[h]
\caption{Cost Budget Tracker}
\KwIn{Slot duration $S$, energy level $C$, polling interval $\Delta t$, EWMA weight $\alpha$}
\KwOut{$E_{\text{usable}}$}

$E_{\text{used}} \leftarrow 0$\;
$P_{\text{EWMA}} \leftarrow 0$\;

\For{$k \leftarrow 1$ \KwTo $\left\lfloor S / \Delta t \right\rfloor$}{
    $E_k \leftarrow \text{energy}(\Delta t)$\;
    $P_k \leftarrow E_k / \Delta t$\tcp*{instantaneous power (W)}
    $P_{\text{EWMA}} \leftarrow \alpha P_k + (1-\alpha) P_{\text{EWMA}}$\;
    $E_{\text{used}} \leftarrow E_{\text{used}} + E_k$\;
    $t_{\text{rem}} \leftarrow S - k \Delta t$\;
    $E^{\text{pred}}_{\text{rem}} \leftarrow P_{\text{EWMA}} \times t_{\text{rem}}$\;
    $E^{\text{pred}}_{\text{tot}} \leftarrow E_{\text{used}} + E^{\text{pred}}_{\text{rem}}$\;
    $E_{\text{usable}} \leftarrow \max\!\bigl(0,\, C - E^{\text{pred}}_{\text{tot}}\bigr)$\;
}
\end{algorithm}

\section{Model Selections}
\label{app:model-selection}
\begin{table}[h]
\caption{Percentage of model selections per task category using JARVIS default model selection method.}

\vskip 0.15in
\centering
\setlength{\tabcolsep}{3pt} 
\footnotesize
\begin{tabular*}{\columnwidth}{@{\extracolsep{\fill}} l l r r r}
\toprule
\multicolumn{1}{l}{Task Type} &
\multicolumn{1}{c}{Model} &
\multicolumn{1}{c}{Chosen} &
\multicolumn{1}{c}{\shortstack[c]{Number\\of likes}} &
\multicolumn{1}{c}{\shortstack[c]{Mentions\\likes}} \\
\midrule

\addlinespace[4pt]
\multicolumn{5}{c}{\textbf{VQA Dataset}} \\
\addlinespace[4pt]

DocVQA & LayoutLM-DQA & 100\% & 174 & 98.07\%\\
\addlinespace[3pt]

IClass & ViT-B16 & 100\% & 169 & 83.19\%\\
\addlinespace[3pt]

\multirow{5}{*}{ICapt}
 & ViT-GPT2 & 78.98\% & 219 & 53.70\%\\
 & BLIP-Capt-L & 13.89\% & 52 & 38.24\%\\
 & BLIP2-6.7B & 4.49\% & 24 & 0\%\\
 & BLIP2-2.7B & 2.09\% & 25 & 0\%\\
 & TrOCR-B-P & 0.55\% & 56 & 0\%\\
\addlinespace[3pt]

OD & DETR-R-50 & 100\% & 129 & 96.05\%\\
\addlinespace[3pt]

VQA & ViLT-B32-VQA & 100\% & 86 & 99.00\%\\

\addlinespace[4pt]
\midrule
\multicolumn{5}{c}{\textbf{OD Dataset}} \\
\addlinespace[4pt]

OD & DETR-R-50 & 100\% & 129 & 99.98\%\\
\addlinespace[3pt]
VQA & ViLT-B32-VQA & 100\% & 86 & 99.86\%\\

\addlinespace[4pt]
\midrule
\multicolumn{5}{c}{\textbf{ICapt Dataset}} \\
\addlinespace[4pt]

ICapt & ViT-GPT2 & 100\% & 219 & 87.21\%\\

\addlinespace[4pt]
\midrule
\multicolumn{5}{c}{\textbf{IGen Dataset}} \\
\addlinespace[4pt]

IGen & SD-1.5 & 100\% & 6367 & 99.83\%\\

\bottomrule
\end{tabular*}
\end{table}

\begin{table}[h]
\caption{Percentage of model selections per task category using Name-Only model selection method.}
\label{app:model-selection-no-likes}
\vskip 0.15in
\vspace*{-0cm}
\centering
\setlength{\tabcolsep}{3pt} 
\footnotesize
\begin{tabular*}{\columnwidth}{@{\extracolsep{\fill}} l l r}
\toprule
\multicolumn{1}{l}{Task Type} &
\multicolumn{1}{c}{Model} &
\multicolumn{1}{c}{Chosen} \\
\midrule

\addlinespace[4pt]
\multicolumn{3}{c}{\textbf{VQA Dataset}} \\
\addlinespace[4pt]

DocVQA & LayoutLM-DQA & 100\%\\
\addlinespace[3pt]

\multirow{2}{*}{IClass}
 & BEiT-B16 & 78.85\%\\
 & DETR-R-50 & 21.15\%\\
\addlinespace[3pt]

\multirow{3}{*}{ICapt}
 & BLIP-Capt-L & 96.66\%\\
 & BLIP2-6.7B & 2.75\%\\
 & TrOCR-B-P & 0.59\%\\
\addlinespace[3pt]

\multirow{3}{*}{OD}
 & DETR-R-101 & 95.37\%\\
 & YOLOv8-S & 3.02\%\\
 & DETR-R-50 & 1.61\%\\
\addlinespace[3pt]

VQA & ViLT-B32-VQA & 100\%\\

\addlinespace[4pt]
\midrule
\multicolumn{3}{c}{\textbf{OD Dataset}} \\
\addlinespace[4pt]

\multirow{2}{*}{OD}
 & DETR-R-101 & 99.82\%\\
 & DETR-R-50 & 0.18\%\\
\addlinespace[3pt]

VQA & BLIP-Capt-L & 100\%\\

\addlinespace[4pt]
\midrule
\multicolumn{3}{c}{\textbf{ICapt Dataset}} \\
\addlinespace[4pt]

\multirow{2}{*}{ICapt}
 & BLIP-Capt-L & 96.79\%\\
 & ViT-GPT2 & 3.21\%\\

\addlinespace[4pt]
\midrule
\multicolumn{3}{c}{\textbf{IGen Dataset}} \\
\addlinespace[4pt]

\multirow{3}{*}{IGen}
 & SD-1.5 & 99.62\%\\
 & OpenJourney & 0.36\%\\
 & SD-2.1 & 0.02\%\\

\bottomrule
\end{tabular*}
\end{table}

\section{Reasoning of Name-Only - OD task} \label{appendix:appendix-OD}
\{"id": "facebook/detr-resnet-101", "reason": "The facebook/detr-resnet-101 model is a robust end-to-end object detection model that utilizes a ResNet-101 backbone, which generally provides better performance and accuracy compared to its ResNet-50 counterpart. It is well-suited for detecting a wide variety of objects in images, making it ideal for the task of object detection. Additionally, it has a local inference endpoint, ensuring faster processing and stability."\}

\section{Reasoning of Name-Only - VQA task} \label{appendix:appendix-VQA}
\{"id": "dandelin/vilt-b32-finetuned-vqa", "reason": "The 'dandelin/ vilt-b32-finetuned-vqa' model is specifically fine-tuned for visual question answering tasks, making it highly suitable for identifying objects in images. It is based on the Vision-and-Language Transformer (ViLT) architecture, which effectively integrates visual and textual information. Additionally, it has a local inference endpoint, ensuring faster response times and greater stability for processing the user's request."\} 

\end{document}